\documentclass[10pt,twocolumn,letterpaper]{article}

\usepackage{iccv}
\usepackage{times}
\usepackage{epsfig}
\usepackage{graphicx}
\usepackage{amsmath}
\usepackage{amssymb}
\usepackage{caption}
\usepackage{booktabs}
\usepackage{subcaption}

\usepackage[pagebackref=true,breaklinks=true,letterpaper=true,colorlinks,bookmarks=false]{hyperref}

\iccvfinalcopy 

\def\iccvPaperID{37} 

\ificcvfinal\fi

\begin{document}

\title{NOVA: NOvel View Augmentation for Neural Composition of Dynamic Objects}

%
\author{Dakshit Agrawal}
\author{Jiajie Xu}

\author{Dakshit Agrawal\thanks{Equal contribution} $^{\text{ 1}}$ \quad \quad \quad Jiajie Xu\footnotemark[1] $^{\text{ 1}}$ \quad \quad\quad  Siva Karthik Mustikovela $^2$ \\ Ioannis Gkioulekas $^1$ \quad \quad \quad Ashish Shrivastava $^2$ \quad \quad \quad Yuning Chai $^2$ \\\vspace{-0.05in} \\ $^1$ Carnegie Mellon University \quad \quad $^2$ Cruise LLC\thanks{Served as independent advisors for the CMU capstone project} \vspace{-0.07in}
}

\maketitle
\ificcvfinal\thispagestyle{empty}\fi

\begin{abstract}
We propose a novel-view augmentation (NOVA) strategy to train NeRFs for photo-realistic 3D composition of dynamic objects in a static scene. Compared to prior work, our framework significantly reduces blending artifacts when inserting multiple dynamic objects into a 3D scene at novel views and times; achieves comparable PSNR without the need for additional ground truth modalities like optical flow; and overall provides ease, flexibility, and scalability in neural composition. Our codebase is on \href{https://github.com/dakshitagrawal/NoVA}{GitHub}.
\end{abstract}


\vspace{-0.25in}
\section{Introduction}
\label{sec:intro}

Photo-realistic composition of objects in a 3D scene has significant applications, one of which is creating realistic content and experiences inside the Metaverse. Despite recent advances in neural radiance fields (NeRFs)~\cite{mildenhall2021nerf}, photo-realistic composition from dynamic monocular videos remains a challenging problem. This is primarily due to the ill-posed nature of this task---multiple scene configurations can lead to identical observed image sequences, a problem we refer to as the 3D structure ambiguity. 

\begin{figure}[h!]
  \centering
   \includegraphics[width=0.9\linewidth]{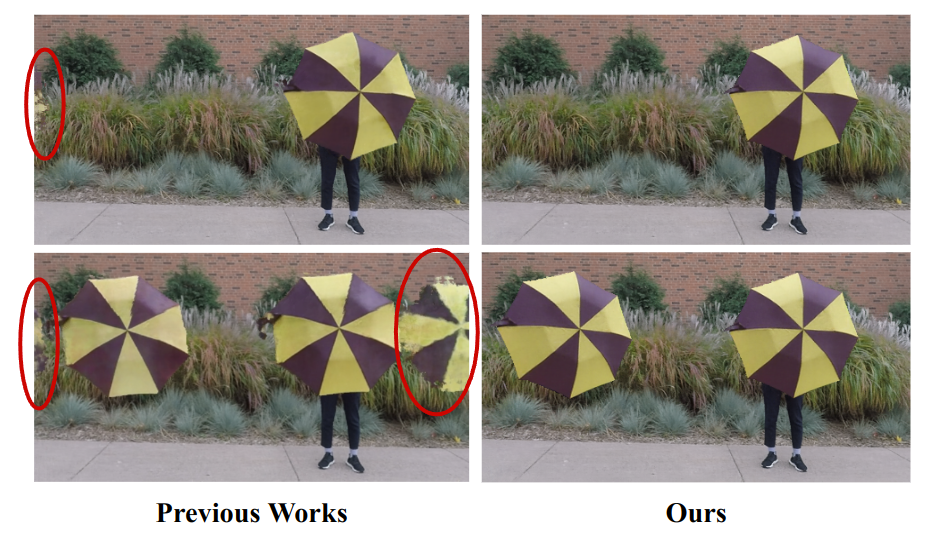}

   \caption{Prior works (left column) have blending artifacts that are amplified when multiple objects are inserted at different points in the same scene. Our method (right column) reduces these blending artifacts significantly.
   \vspace{-0.18in}}
   \label{fig:artifacts}
\end{figure}

Current approaches for this task \cite{gao2021dynamic,li2021neural} build implicit representations of the static scene and dynamic objects separately by predicting a per-point blending factor along with color and density. To deal with structure ambiguity, these methods also predict modalities such as 3D scene flow and depth to regularize the prediction within each frame and between neighboring frames. This requires ground truth data for these modalities, thus limiting applicability. These approaches also suffer from blending mask prediction errors when rendering a novel view, causing blending artifacts at the boundaries of the image that are not present in the reference frustum. This effect is amplified when inserting multiple objects into the scene and dramatically degrades the rendering quality (see Fig. \ref{fig:artifacts}). 

We introduce a framework, NOVA, that helps mitigate these issues. NOVA reduces blending artifacts by augmenting NeRF with losses for different views during training and requiring the network to predict consistent masks and colors across novel views. NOVA additionally extends prior works to facilitate learning different dynamic objects of the scene using separate implicit representations and controlling their movement by manipulating these representations. NOVA does not require 3D scene flow regularization, thus removing the need for a scene flow predictor during data preparation and reducing training time without impacting PSNR. In summary, our contributions are three-fold:
\begin{enumerate}
    \item a flexible NeRF composition framework to add an arbitrary number of dynamic objects into a static 3D scene;
    \item a novel-view augmentation strategy for learning better per-point blending factors;
    \item corresponding novel-view losses for high rendered image fidelity.
\end{enumerate}

\section{Related Work}
\label{sec:related}

\begin{figure*}[h!]
  \centering
   \includegraphics[width=0.9\linewidth]{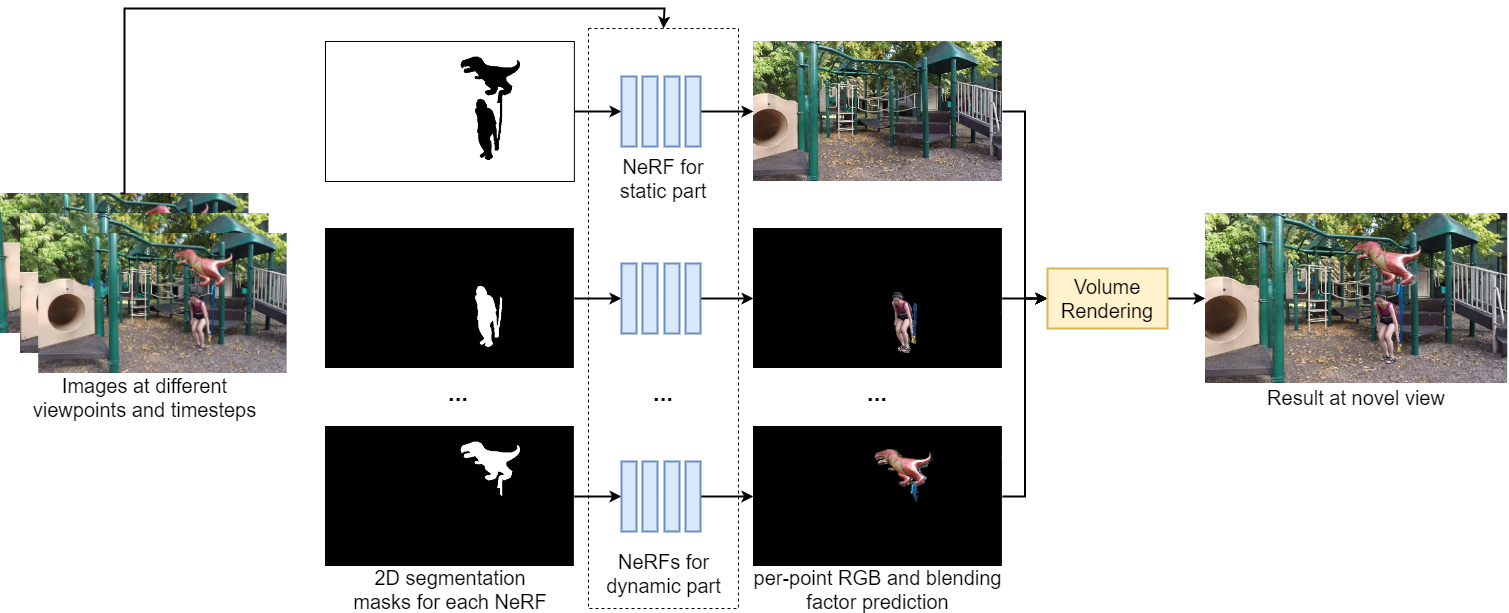}

   \caption{Overview of our training framework. Based on the 2D segmentation masks, separate NeRFs are initialized. These NeRFs predict per-point RGB color and blending factors, which are passed through a differentiable volume renderer to generate the final composed image from a novel viewpoint.
   \vspace{-0.15in}
   }
   \label{fig:nova_arch}
\end{figure*}

\paragraph{Object composition via inverse-rendering.}
Inserting objects into a scene requires properties like lighting, depth, geometry, and material. \cite{GAO_2019, Sengupta_2019, Li_2020, Zhu_2022} estimate these properties for an indoor scene from a single image. For outdoor scenes, a high dynamic range light field is necessary to represent sun and sky \cite{Hold_Geoffroy_2019, Wang_2022}, and adversarial methods are commonly used to train photo-realistic results \cite{Wang_2022, Kim_2021, Wang_2019, Niemeyer2020GIRAFFE}.

\paragraph{Composing dynamic objects using NeRFs.}
NeRFs \cite{mildenhall2021nerf} achieve impressive novel-view synthesis results with a simple formulation for static scenes, encouraging research to compose multiple NeRFs. Guo \etal \cite{guo2022objectcentric} proposed training per-object scattering functions for proper lighting effects during composition. Yang \etal \cite{yang2021objectnerf} separated the scene into background and object branches, using 2D segmentation as supervision. To allow for 3D pose control, Ost \etal \cite{Ost_2021_CVPR} proposed a learnable scene graph to decompose dynamic objects into nodes encoding transformation and radiance. Tancik \etal \cite{tancik2022blocknerf} proposed a framework to tune and compose individually trained NeRFs into city-scale scenes.
\paragraph{Novel-view synthesis for dynamic videos.}
Current works either learn a static canonical radiance field, with a second per-time-step field to apply deformation \cite{tretschk2021nonrigid, park2021nerfies, pumarola2020d}, or learn a dynamic radiance field directly conditioned on time \cite{li2021neural, XianHK021, du2021nerflow, gao2021dynamic, Tian_2022_Mononerf}. For the latter direction, it is common to learn a scene flow field \cite{Vedula_1999} concurrently and constrain adjacent frames for pixel consistency. Besides scene flow, Li \etal \cite{li2021neural} also applied geometric consistency and depth as prior; Gao \etal \cite{gao2021dynamic} introduced additional auxiliary losses. Tian \etal \cite{Tian_2022_Mononerf} propose a flow-based feature aggregation module to incorporate spatial and temporal features.


\section{Method}
\label{sec:method}

Our framework is inspired by Gao \etal \cite{gao2021dynamic}, which jointly trains two NeRFs that separately handle the time-invariant static and time-varying dynamic parts of a monocular video. The static NeRF predicts the per-point color and density $(c, \sigma)$ given the point's position and viewing direction $(x,y,z,\theta,\phi)$. The dynamic NeRF predicts the per-point color, density, scene flow, and blending factor $(c, \sigma, s_f,s_b, \beta)$ given the point's position, viewing direction, and time $(x,y,z,\theta,\phi, t)$. Ground-truth optical flow is used to learn the scene flow, and several regularizing losses are applied to scene flow and depth to resolve the 3D structure ambiguity when learning from a monocular view. The NeRF composition is done in an unsupervised manner using the per-point blending factors $\beta$.

This approach works well for scene reconstruction but produces blending artifacts when manipulating the scene (see Fig. \ref{fig:artifacts}). We introduce a framework with three novel modules to alleviate these issues, described in detail in the subsequent sections. We also remove the losses based on ground-truth optical flow in Gao \etal \cite{gao2021dynamic} from our framework to reduce the amount of supervision.

\subsection{Multiple NeRFs}

Our framework uses separate NeRFs to learn different parts of the scene. Each NeRF is provided a segmentaion mask of the scene and is either static or dynamic based on the dynamicity of the scene parts it models (see Fig. \ref{fig:nova_arch}). The static and dynamic NeRF architectures are similar to that of Gao \etal \cite{gao2021dynamic}. The final RGB image is produced from a novel viewpoint by combining the outputs of all the NeRFs as follows:
\begin{equation}
\label{eq:colfull}
   \mathbf{C}^{full}_P (\mathbf{r}) = \sum_{k=1}^K T_k^{full} \left( \sum_{n=1}^{num\_NeRFs} \alpha^n_{k} \beta^n_{k} \mathbf{c}^n_k \right) 
\end{equation}

where $K$ is the number of samples along the ray $\mathbf{r}$, $T_k^{full}$ is the transmittance at the $k^{th}$ sample along the ray after accounting for rays from all the NeRFs, and $\alpha^n_k$, $\beta^n_k$, and $\mathbf{c}^n_k$ are the alpha, blending factor, and color respectively predicted by the $n^{th}$ NeRF for the $k^{th}$ sample along the ray.

\subsection{Novel-View Augmentation}
\label{sec:novaug}

\begin{figure}[h!]
  \centering
   \includegraphics[width=0.90\linewidth]{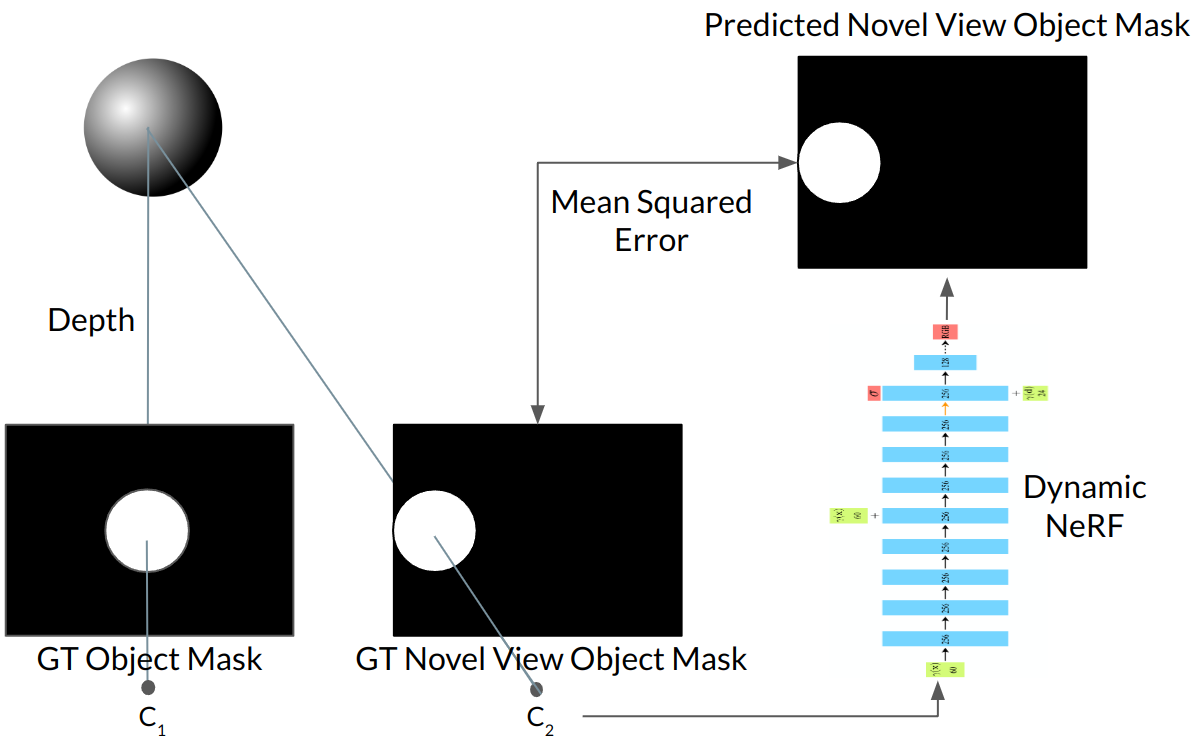}

   \caption{Novel-view augmentation training strategy 
   }
   \label{fig:novaug}
\end{figure}

Our novel-view augmentation training strategy reduces blending artifacts when manipulating multiple dynamic objects and composing them into the scene. During training, we shift the camera responsible for the dynamic object to a novel view (see Fig. \ref{fig:novaug}). Given the camera's relative transformation, we calculate the ground truth segmentation mask at the novel view using stereo geometry. Points are sampled along the rays of the camera at the novel viewpoint $C_2$ and passed through the corresponding NeRF. We render the predicted segmentation mask $\mathbf{M}^n_P$ for the $n^{th}$ NeRF as follows:
\begin{equation}
    \mathbf{M}^n_{P} (\mathbf{r}) = \sum_{k=1}^K T_k^{full} \alpha_k^{full} \beta^n_k
\end{equation}
where $T_k^{full}$ is the transmittance and $\alpha_k^{full}$ is the alpha at the $k^{th}$ sample along the ray after accounting for rays from all the NeRFs, and $\beta^n_k$ is the blending factor predicted by the $n^{th}$ NeRF at the $k^{th}$ sample along the ray. This augmentation strategy can be applied to other ground truths available for training like RGB images.

\begin{table*}[h!]
  \centering
  \begin{tabular}{l|c|c|c|c|c|c|c|c}
    Method & Balloon1 & Balloon2 & Jumping & Playground & Skating & Truck & Umbrella & Average  \\
    \midrule
    NeRF + time &  17.32 & 19.66 & 16.72 & 13.79 & 19.23 & 15.46 & 17.17 & 17.05\\
    Yoon \etal \cite{yoon2020novel} &  18.74 & 19.88 & 20.15 & 15.08 & 21.75 & 21.53 & 20.35 & 19.64\\
    Li \etal \cite{li2021neural} & 21.35 & 24.02 & \textcolor{red}{24.10} & 20.85 & \textcolor{blue}{\underline{28.88}} & \textcolor{blue}{\underline{23.33}} & 22.56 & \textcolor{blue}{\underline{23.58}}\\
    Gao \etal \cite{gao2021dynamic} & \textcolor{blue}{\underline{21.43}} & \textcolor{red}{26.59} & \textcolor{blue}{\underline{23.57}} & \textcolor{red}{23.74} & \textcolor{red}{31.92} & \textcolor{red}{25.50} & \textcolor{blue}{\underline{22.68}} & \textcolor{red}{25.06} \\
    \midrule
    Ours & \textcolor{red}{21.52} & \textcolor{blue}{\underline{25.08}} & 20.27 & \textcolor{blue}{\underline{22.31}} & 27.73 & 23.31 & \textcolor{red}{23.08} & 23.33\\
  \end{tabular}
  \caption{We compare PSNR of our method against other methods that report their PSNR on Dynamic Scene Dataset \cite{yoon2020novel}. The best results are highlighted in \textcolor{red}{red} while the second best are in \textcolor{blue}{\underline{blue}}. Our model performs comparably to other methods despite not using ground-truth optical flow supervision.}
  \label{tab:quan}
\end{table*}

\subsection{Novel-View Losses}
\label{sec:novloss}
    
We introduce a few losses to ensure high image fidelity when placing objects at novel points in the scene.

\noindent \textbf{Novel-View Mask Loss.} We take the squared error loss between the predicted and ground-truth masks for the novel viewpoint:
\begin{equation}
    \mathcal{L}_{nvm} = \sum_{n=1}^{num\_NeRFs} \sum_{ij} \| \mathbf{M}^{n}_{GT} (\mathbf{r}_{ij}) - \mathbf{M}^{n}_{P} (\mathbf{r}_{ij})\|_2
\end{equation}

\noindent \textbf{Per-Camera Novel-View RGB Loss.} We render the RGB image of each NeRF as follows:
\begin{equation}
    \mathbf{C}^n_{P} (\mathbf{r}) = \sum_{k=1}^K T^n_k \alpha^n_k \beta^n_k \mathbf{c}^n_k
\end{equation}

We take the squared error loss between the predicted and the ground-truth RGB image from the novel viewpoint of only the pixels for which the NeRF is responsible for:
\begin{equation}
    \mathcal{L}_{nvcn} = \sum_{n=1}^{num\_NeRFs} \sum_{ij} \mathbf{M}_{GT}^n (\mathbf{r}_{ij}) \| \mathbf{C}_{GT} (\mathbf{r}_{ij}) - \mathbf{C}^{n}_{P} (\mathbf{r}_{ij})\|_2
\end{equation}

\noindent \textbf{Full Novel-View RGB Loss.} After rendering the final RGB image using Eq. \ref{eq:colfull}, we take the squared error loss with the ground truth full RGB image as follows:
\begin{equation}
    \mathcal{L}_{nvcf} = \sum_{ij} \| \mathbf{C}_{GT} (\mathbf{r}_{ij}) - \mathbf{C}^{full}_{P} (\mathbf{r}_{ij})\|_2
\end{equation}

\noindent \textbf{Blending Loss.} To ensure the contributions of all the NeRFs for a particular point sum to one, we introduce a blending loss:
\begin{equation}
\mathcal{L}_{nvb} = \sum_{ijk} \left| \left(\sum_{n=1}^{num\_NeRFs} \beta^n_{ijk} \right) - 1 \right|
\end{equation}

\noindent \textbf{Alpha Loss.} We force the NeRFs to not predict anything outside the masks they are responsible for by explicitly adding a loss for alphas to be 0 outside the camera mask:
\begin{equation}
 \mathcal{L}_{nva} = \sum_{n=1}^{num\_NeRFs} \sum_{ij} \left(1-\mathbf{M}^n_{GT} (\mathbf{r}_{ij})\right) \cdot \left( \sum_k \left| \alpha^n_{ijk} \right| \right)   
\end{equation}

\begin{figure*}[h!]
  \centering
   \includegraphics[width=0.92\linewidth]{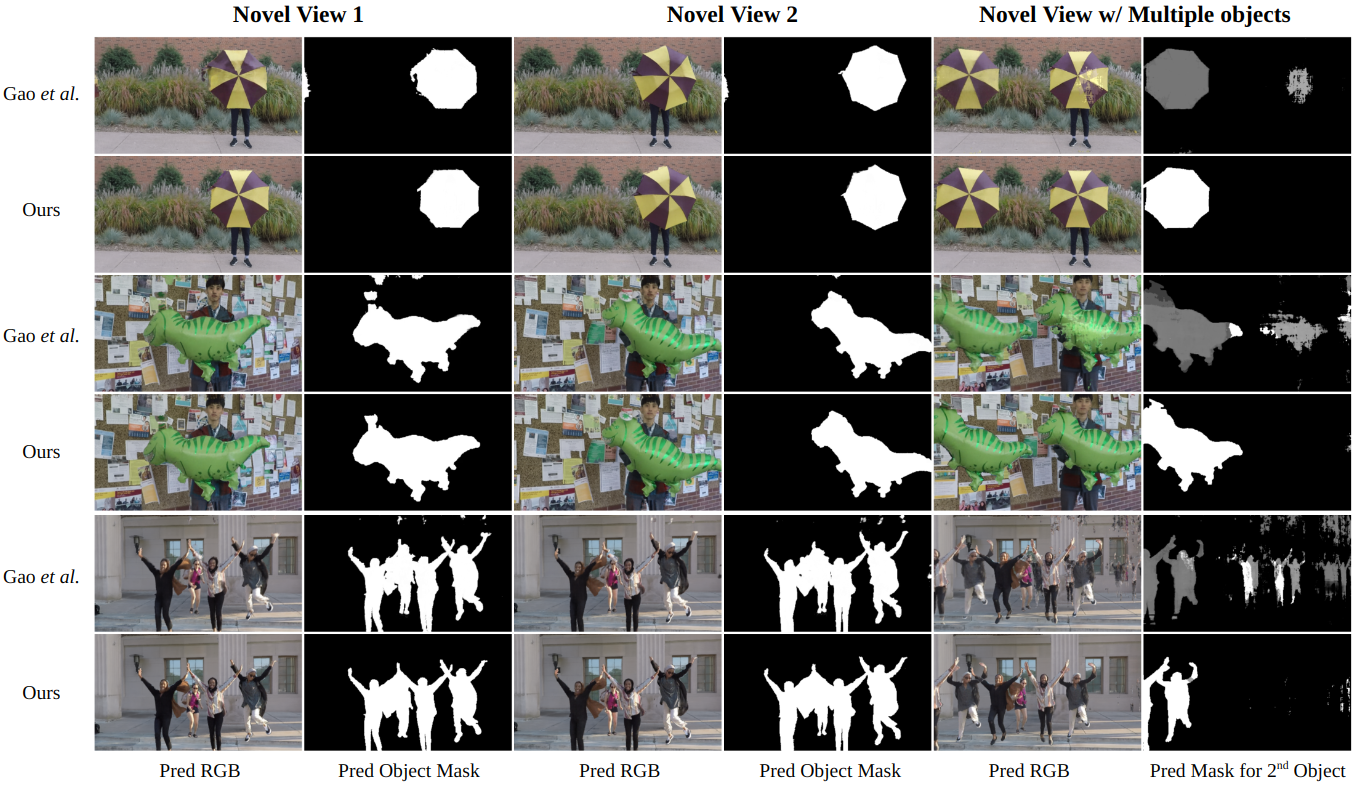}

   \caption{Qualitative results of our model on Umbrella, Balloon1, and Jumping scenes. Compared with Gao \etal \cite{gao2021dynamic}, our novel-view augmentation training significantly reduces artifacts in the novel-view mask prediction, and produces images with higher fidelity, especially when composing multiple objects in a scene.
   \vspace{-0.05in}
   }
   \label{fig:qual}
\end{figure*}

\section{Experimental Results}
\subsection{Dataset}
\label{sec:dataset}

We use the preprocessed Dynamic Scene Dataset \cite{yoon2020novel} provided by Gao \etal \cite{gao2021dynamic}, which contains video sequences for seven scenes, each consisting of a static background and moving objects. Each sequence has 12 images captured at different time steps and camera poses, which make them effectively monocular.

\subsection{Evaluation}
\label{sec:eval}

\subsubsection{Quantitative Evaluation}

We evaluate the image fidelity quantitatively by assessing the PSNR between the synthesized image and the corresponding ground truth image at a fixed viewpoint but changing time. Our framework performs comparably to other methods without the need for additional modalities of ground truth data like optical flow (see Tab. \ref{tab:quan}).

\subsubsection{Qualitative Evaluation}

We compare our novel-view renderings with Gao \etal \cite{gao2021dynamic} in Fig. \ref{fig:qual}. Our framework reduces blending artifacts, as visible clearly from our predicted object masks and generated final images, with the improvement being significant when composing multiple dynamic objects.

\subsubsection{Ablation Study}

\vspace{-0.1in}
\begin{figure}[h!]
  \centering
   \includegraphics[width=\linewidth]{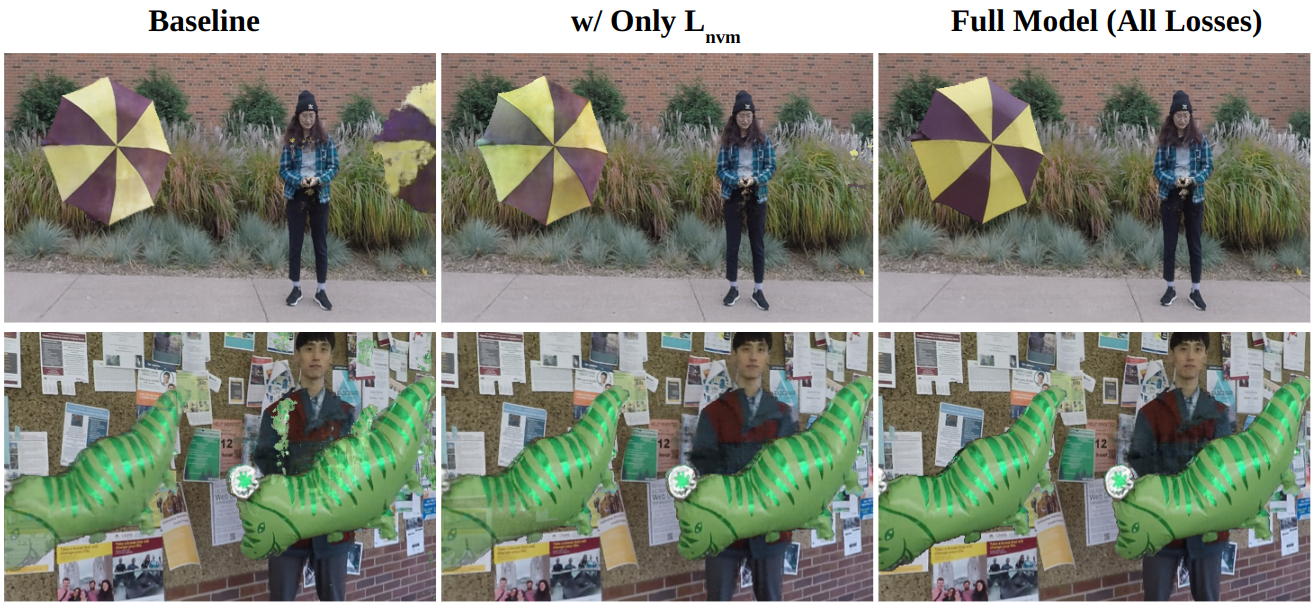}

   \caption{Ablation study on $\mathcal{L}_{nvm}$ and novel-view RGB losses.\vspace{-0.1in}}
   \label{fig:ablation}
\end{figure}

We study the impact of each of our losses on the quality of the final image. As seen in Fig. \ref{fig:ablation}, using just $\mathcal{L}_{nvm}$ can remove the blending artifacts, but RGB losses are necessary to ensure the inserted objects have proper color. 






\section{Conclusion}
\label{sec:conclusion}

We have introduced a framework, NOVA, for the neural composition of dynamic scenes using NeRFs. Our major contributions are three modules: multiple NeRFs, novel view augmentation, and novel view losses. Using monocular dynamic video, object segmentation masks, and depth information, our results demonstrate our framework's reliability, ease, flexibility, and scalability of inserting multiple dynamic objects into a scene photo-realistically.

{\small
\bibliographystyle{ieee_fullname}
\bibliography{egbib}
}

\end{document}